\documentclass{article}
\usepackage{spconf} 
\usepackage{amsmath,graphicx}
\usepackage{url}

\usepackage{epstopdf}
\usepackage{subfig}
\usepackage{placeins}

\title{Automatic breast cancer grading in lymph nodes using\\ a deep neural network}

\name{T. Wollmann, K. Rohr}
\address{University of Heidelberg, BioQuant, IPMB, and DKFZ Heidelberg,\\
Dept. Bioinformatics and Functional Genomics, Biomedical Computer Vision Group,\\
Im Neuenheimer Feld 267, 69120 Heidelberg, Germany}

\begin{document}
%
\maketitle
\begin{abstract}
The progression of breast cancer can be quantified in lymph node whole-slide images (WSIs). We describe a novel method for effectively performing classification of whole-slide images and patient level breast cancer grading. Our method utilises a deep neural network. The method performs classification on small patches and uses model averaging for boosting. In the first step, region of interest patches are determined and cropped automatically by color thresholding and then classified by the deep neural network. The classification results are used to determine a slide level class and for further aggregation to predict a patient level grade. Fast processing speed of our method enables high throughput image analysis.
\end{abstract}
\section{Introduction}
\label{sec:intro}
Breast cancer is the most mortal cancer disease for women, worldwide \cite{mcguire2015effects}. The progression of the disease is quantified by pathologists using whole-slide images (WSIs) of lymph nodes, which are stained with hematoxylin and eosin (H\&E). For grading the progression of cancer diseases the TNM system is used \cite{edge2010american}. In the TNM classification system, the parameter N describes the degree of cancer spread to regional lymph nodes. Currently, pathologists perform the pathologic N-stage grading manually, which is tedious, time-consuming, and error prone. Automation of this process could lead to the reduction of manual work and errors.

In the past years several challenges for whole-slide image analysis were conducted (e.g., \cite{tupac:2016:Online, cam:2016:Online}). However, most methods use dense classification of the whole-slide images, which is slow and requires processing of many non-meaningful image regions \cite{wang2016deep, chen2016identifying, liu2017detecting}. Sparse classification methods speed up the classification process \cite{wollmann2017automatic,wollmann2017hough}. Thus, such methods could run on the workstation of a pathologist and support the decision process. However, sparse classification generally reduces the performance which, however, can be alleviated by using model averaging. 

In this work, we describe a method which uses a deep neural network (DNN) for sparse classification of WSIs. Our approach determines regions of interest (ROIs) within the WSIs using color thresholding and morphological operations followed by sparse classification using a deep neural network. The network is trained and tested using the same image resolution. Since the WSIs have varying downsampling factors, we employ a normalization.

\section{Method}
\label{sec:methods}
Our method uses a region of interest selection method and a deep neural network to perform sparse classification of the WSI. The individual classification results are aggregated to a slide wide class. Additional decision rules, provided by the CAMELYON17 team, are used for patient level classification. 

\begin{figure}[!ht]
   \center
   \includegraphics[width=0.45\textwidth]{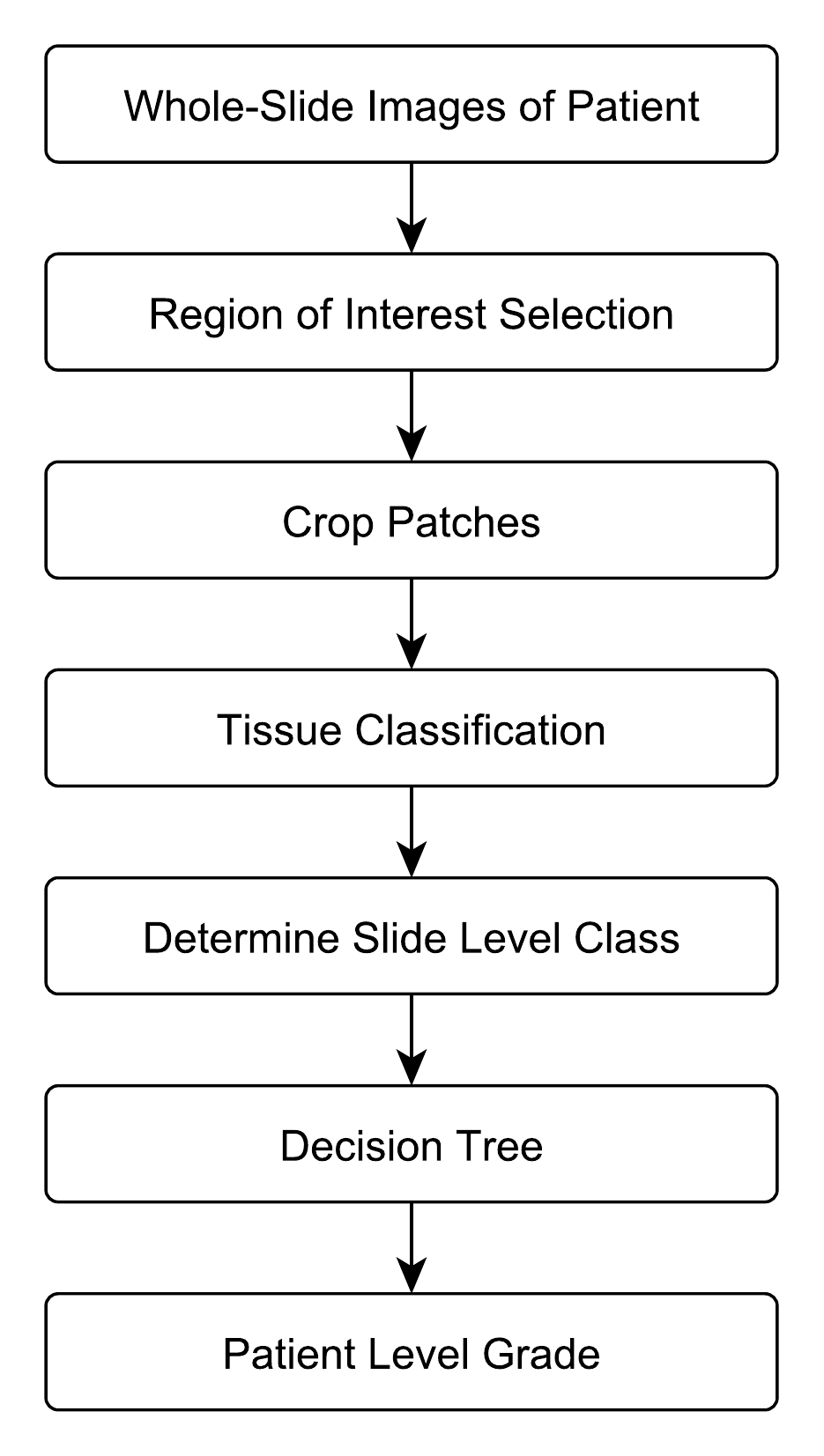}
   \caption{Overall workflow of our method}
   \label{fig:workflow}
\end{figure}
Figure \ref{fig:workflow} illustrates the overall workflow of our method. The deep neural network is implemented in Tensorflow \cite{abadi2016tensorflow}.

\subsection{Region of Interest Selection}
A region of interest (ROI) is determined by color thresholding similarly to the method in \cite{wollmann2017automatic}. Colour thresholding is performed using a ratio of the intensities of 0.9 in the green and red channel. Afterwards, a median filter with a 50 pixels disk shaped structuring element is applied. Based on the whole slide with a downsampling factor of 64 as input, a region of interest map is calculated.
\begin{figure}[!ht]
	\centering
    \includegraphics[width=0.45\textwidth]{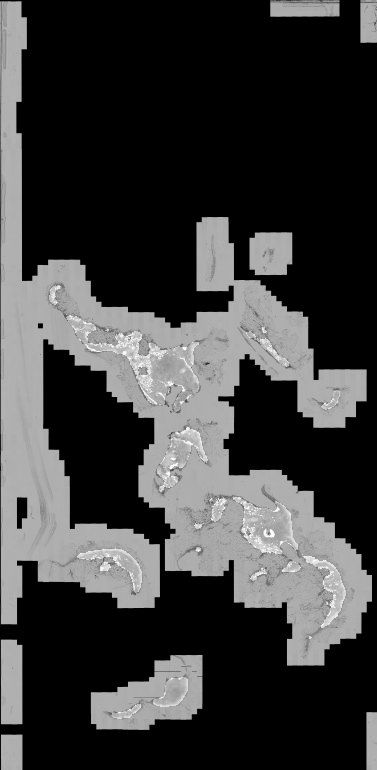}
    \caption{Example demonstrating the region of interest selection mechanism}
    \label{fig:attention}
\end{figure}
Figure \ref{fig:attention} shows examples of regions of interest determined by the region of interest selection method.
 
For tissue classification after ROI selection, 20 region centroids with a size of $512\times512$ pixels with a downsampling factor of 64 are sampled from the WSI using the region of interest map.

\subsection{Tissue Classification}
\begin{figure}[!ht]
	\centering
    \includegraphics[width=0.45\textwidth]{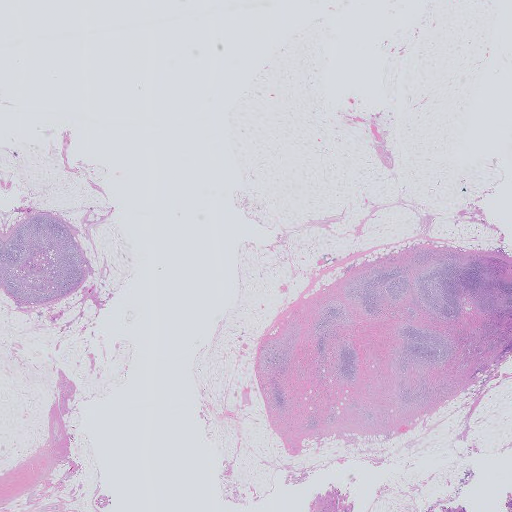}
    \caption{Example $512\times512$ pixels image patch}
    \label{fig:patch}
\end{figure}

Tissue classification is performed on image regions determined by the region of interest selection mechanism. Computed centroid coordinates are used to extract $512\times512$ pixels image patches with a downsampling factor of 64. A sample patch is show in Figure \ref{fig:patch}. The patches are augmented using random rotations and flipping, and used for model averaging. All patches are classified into a one-hot encoding of the four classes: contains isolated tumour cells (ITC), contains a macro metastasis, contains a micro metastasis or is negative. 

The deep neural network is inspired by densely connected deep neural networks \cite{huang2016densely}. In the first block a $3\times3$ convolution is performed yielding 32 feature maps. Since it was empirically observed that the first layers benefit from negative parts of activation functions we replaced the original ReLU with PReLU \cite{he2015delving, paszke2016enet}. For downsampling and feature disentangling we alternate densely connected blocks with downsampling blocks as described in \cite{huang2016densely}. Within the densely connected blocks we use six densely connected feature map extractors. Each downsampling block doubles the number of feature maps. The predicted class is calculated using global average pooling \cite{lin2013network}, a fully connected layer with 128 hidden units, and the softmax operator. All non-linear layers are initialised using MSRA initialisation and we used ReLU activation functions \cite{he2015delving}.
            
\subsection{Patient Level Grading}
A slide level class is calculated by adding the activations of the class activations of the last layer and calculating the maximum over all 20 patches. By selecting the maximum after adding up the activations of multiple predictions we perform model averaging. Instead of using the maximum, we also tried to rank the classes by macro metastasis, micro metastasis, ITC, and negative, and picked the highest ranked class. However, this technique did not improve the results on the validation set.

After determining slide wide classes a patient level grade is determined using the following decision rules:
\begin{itemize}
\item \textbf{pN0:} No micro-metastases or macro-metastases or ITCs found.
\item \textbf{pN0(i+):} Only ITCs found.
\item \textbf{pN1mi:} Micro-metastases found, but no macro-metastases found.
\item \textbf{pN1:} Metastases found in 1-3 lymph nodes, of which at least one is a macro-metastasis.
\item \textbf{pN2:} Metastases found in 4-9 lymph nodes, of which at least one is a macro-metastasis.
\end{itemize}

\subsection{Model Training}
The deep neural network model is trained using the cross entropy loss. We trained the network for six epoch within eight hours on 80\% of the data and kept 20\% for validation of the model. In each epoch the training data class occurrence frequencies are balanced and the image patches are augmented.

The tissue classification network is trained using mini batches with 10 image patches each. The $512\times512$ pixels image patches are extracted from the original WSIs by a downsampling factor of 64 using our color thresholding method, augmented using random rotation, flipping, random color shifts, and elastic deformation, and passed to the deep neural network. Data augmentation and transfer to the GPU node is optimised using multi threaded data streaming jobs running on the CPU.

The models are trained using the Adam optimizer \cite{kingma2014adam} with an initial learning rate $l_{init}=0.001$, as well as $\beta_1=0.9$ and $\beta_2=0.999$.

\subsection{Experimental}
We measured the speed of our algorithm on the challenge dataset. The classification of a $512\times512$ pixels patch requires on average 0.04 seconds on an Intel i7-6700K workstation with a NVIDIA Geforce GTX 1070. We measured a total computation time of 0.78 seconds for a WSI and 3.90 seconds for predicting a patient level grade.

\section{Discussion and Conclusion}
\label{sec:conclusion}
We presented an automatic approach for patient level breast cancer grading for lymph node WSIs. Our approach utilises a region of interest selection mechanism and a densely connected deep neural network to perform sparse classification. The sparse classification results are aggregated using decision rules. The algorithm determines a patient level grade based on five WSIs in about 4 seconds on a standard workstation by using a region of interest selection mechanism and classifying big image crops. A trade-off between statistical power and speed can be achieved natively by changing the number of patches used for the sparse classification. Our work contributes to the field of breast cancer grading by providing a fast method, which can be trained using only slide level annotations.
\\\\
\noindent \textbf{Acknowledgements}
\label{sec:ack}

\noindent This work was supported by the BMBF-funded Heidelberg Center for Human Bioinformatics (HD-HuB) within the German Network for Bioinformatics Infrastructure (de.NBI) \#031A537C.

\bibliographystyle{ieeetr}
\small
\bibliography{refs}

\end{document}